\documentclass[preprint,12pt,authoryear]{elsarticle}

\usepackage{amssymb}
\usepackage{amsmath}
\usepackage{placeins}
\usepackage{float}
\usepackage{url}
\usepackage[normalem]{ulem}

\journal{Remote Sensing of the Environment}

\begin{document}

\begin{frontmatter}

\title{Real-time physics inversion for retrieval of sub-pixel wildfire temperatures from VSWIR imaging spectroscopy}

\author[jpl]{William R. Keely}

\author[jpl]{Philip G. Brodrick}

\author[utah]{Katherine Mistick}

\author[jpl]{Adam Chlus}

\author[jpl]{Robert O. Green}

\author[utah]{Philip E. Dennison} 

\affiliation[jpl]{organization={Jet Propulsion Laboratory, California Institute of Technology},
            state={CA},
            country={United States of America}}

\affiliation[utah]{organization={School of Environment, Society \& Sustainability, University of Utah},
            city={Salt Lake City},
            state={UT},
            country={United States of America}}

\begin{abstract}
In this work, we present a wildfire temperature retrieval framework for VSWIR imaging spectroscopy data, employed on data from NASA's Airborne Visible Infrared Imaging Spectrometer (AVIRIS-3). The retrieval framework utilizes a full-physics approach in which a forward model is employed to resolve both solar and emitted radiance derived from a temperature distribution and utilizes the full spectral range in the residual fit. To optimize the forward model retrieval, we use state-of-the-art nonlinear least squares methods implemented for fast convergence on the on-board GPU, allowing for estimation of effective fire temperature within flight cadence. We verify the forward model assumptions on simulated spectra with an injected thermal signature and find good agreement with an RMSE of $41.8$ Kelvin (K). We apply the retrieval over the full 2025 FireSense AVIRIS-3 campaign, totaling 168 overflights with probable active fire spectra, and demonstrate a residual radiance fit of $\leq 10\%$ across bands in the short-wave infrared (SWIR). Lastly, we verify the applicability of the retrieved posterior fire temperature parameters to generalize to space-borne imaging spectrometers such as EMIT, by retrieving at coarsened spatial resolution. We find that the posterior distribution exhibits good coverage of the underlying sub-pixel temperature range with an absolute error of $30$ K across quantiles and a mean absolute error of $27.16$ K between spatial resolutions.  
\end{abstract}


\end{frontmatter}

\section{Introduction}
\label{introduction}

Wildfire is one of the most significant short-timescale forcings on
terrestrial ecosystems, atmospheric composition, and human safety in
fire-prone regions. The combustion temperature of a fire, its areal
extent, and the distribution of flaming
and smoldering phases together govern the emission of trace gasses
and particulates, the depth of soil heating, and the post-fire
trajectory of ecosystem recovery
\citep{wooster2005,kaufman1998,giglio2003}. Accurately estimating
fine-scale fire temperature from spaceborne and airborne measurements
is therefore a long-standing priority for emissions inventories,
fire-weather modeling, and operational incident response.

Remote sensing fire temperature retrievals dominantly rely on modeling at-sensor radiance as a mixture of fire-emitted and background radiances. \citet{dozier1981} developed a “bispectral” method applied to Advanced Very High Resolution Radiometer (AVHRR) data that modeled mid infrared (MIR) and thermal infrared (TIR) brightness temperatures as the sum of fire-emitted and background-emitted endmembers, multiplied by fractional area for each endmember. Subsequent work examined similar spectral mixing models applied to Moderate Resolution Imaging Spectrometer (MODIS) data \citep[e.g.,][]{eckmann2008,giglio2014}. Limitations of MIR-TIR coarse resolution mixing model retrievals of fire temperature include high sensitivity to estimated background temperature \citep{giglio2001,giglio2014} and the assumption of a single fire temperature at scales of hundreds of meters. 

The shortwave infrared (SWIR) spectral region (1400-2500 nm) features strong emitted radiance from combustion temperatures higher than 500 K, negligible background thermal radiance, and (relative to hot fires) weak solar reflected radiance \citep{dennison2006}. Flaming combustion temperatures ($\sim$850-1500 K) add emitted radiance to near infrared (NIR) wavelengths (700-1400 nm), and for temperatures higher than 1150 K, peak emission occurs in the SWIR. These emission characteristics make imaging spectrometers operating in the VSWIR (visible-to-shortwave infrared, $\sim$400-2500 nm) well-suited to fire temperature estimation. Contiguous, narrow-band sampling across the VSWIR spectral region provides hundreds of discrete bands for resolving the magnitude and shape of fire-emitted radiance while accounting for background reflected solar radiance and atmospheric effects. 

Imaging spectrometer fire temperature retrievals have also dominantly relied on spectral mixing models. \citet{dennison2006} demonstrated modeling the at-sensor radiance spectrum as the best fit mixture of a single-temperature fire-emitted radiance endmember and a background reflected solar radiance endmember. They applied their model to 5 m spatial resolution Airborne Visible Infrared Imaging Spectrometer (AVIRIS) data acquired over wildfire. \citet{waigl2019} and \citet{amici2022} extended this spectral mixing framework to include two fire temperature and two background endmembers, applying their models to 30 m EO-1 Hyperion and PRISMA (PRecursore IperSpettrale della Missione Applicativa) data, respectively. Using AVIRIS and MASTER (MODIS/ASTER Airborne Simulator) data spanning the visible through TIR, \citet{dennison2011comparison} established that SWIR bands are essential for consistent temperature and fractional-area modeling above 800 K. 

The present work is motivated by several important limitations of spectral mixture modeling applied to imaging spectrometer data for fire temperature estimation. First, previous models have assumed one or two single-temperature endmembers. While this assumption is more reasonable at fine spatial resolution, in situ measurements have demonstrated that combustion temperatures vary at millimeter scales \citep{wotton2012}. Ideally, temperature estimation should characterize pixels as possessing a range of fire temperatures contributing to emission. Second, atmospheric effects have been oversimplified in previous modeling. Previous approaches simplify fire-emitted radiance using radiative transfer, but have not accounted for varying path length at the pixel scale (e.g., elevation, view zenith angle). This can result in SWIR water vapor and carbon dioxide residuals that influence temperature retrieval. Third, saturated bands have been inconsistently handled by past methods, with some algorithms utilizing all unsaturated bands as determined on a per-pixel basis \citep{dennison2006,dennison2011comparison} while others avoided saturated spectra or frequently saturated bands \citep{amici2022,matheson2012,waigl2019}. Fourth, \citet{matheson2012} showed that prior spectral-mixture fire temperature retrievals are sensitive to the instrument spatial resolution, with modeled fire temperatures biased cooler as fine airborne spectra are aggregated to coarser footprints, a sensitivity that any retrieval intended for both airborne and spaceborne deployment must contend with.

In this work we present a forward model inversion for wildfire
temperature estimation from VSWIR imaging spectroscopy that
addresses each of these limitations. A 6S-based radiative
transfer parameterization \citep{vermote1997} is coupled into the
inversion through pre-solved water-vapor look-up tables built
using the strategy of \citet{thompson2018isofit} and
\citet{brodrick2021srtmnet}, removing  residual atmospheric
structure. The sub-pixel
temperature distribution is parameterized internally as a
two-component Gaussian mixture with a flaming and a smoldering
component, with each fit using a different SWIR window. This decouples
the inversion into temperature regimes that the fit weights
dynamically, giving the retrieval flexibility across the range of
fire conditions encountered in a scene. The same structure
provides robustness to saturation, since saturated bands are
masked from their respective windows on a per-pixel basis and the
loss of constraint is contained to the affected regime rather
than collapsing the fit globally. The mixture is collapsed to a
single moment-matched Gaussian per pixel, summarizing the range
of sub-pixel temperatures consistent with the observed spectrum
as an effective fire temperature $\mathbb{E}[T]$ and a sub-pixel
temperature spread $\sigma[T]$. We apply this retrieval to all 168 unique
AVIRIS-3 flight passes acquired during the 2025 FireSense campaign, verify the forward model on simulated
spectra with known injected temperatures, demonstrate retrieval
consistency across spatial resolutions by aggregating fine
($\sim$5 m) AVIRIS-3 pixels to a coarse ($\sim$60 m) grid that
emulates a spaceborne footprint,
and characterize the campaign-scale distribution of retrieved
effective fire temperatures. The inversion is implemented to run on the AVIRIS-3
on-board GPU, enabling real-time fire temperature products. Coarsened data demonstrate the applicability of this new method for current and future spaceborne imaging spectrometers, including EMIT 
\citep{green2020emit} and the upcoming
EAGLE-VSWIR mission \citep{thompson2026next}.

\section{Methods}
\label{methods}
The retrieval framework comprises a per-pixel radiative-transfer forward
model with explicit solar and thermal components, inverted by a
two-stage Adam optimization of a windowed least-squares loss
between modeled and observed radiance. Atmospheric water vapor is
pre-solved per pixel and held fixed during the temperature
inversion, with the atmospheric look-up table cached per scene.
The sub-pixel temperature distribution is parameterized as a
two-component Gaussian mixture during inversion and reported as a
moment-matched single Gaussian whose mean and standard deviation
summarize the range of sub-pixel temperatures consistent with the
observed spectrum; we refer to the mean $\mathbb{E}[T]$ as the
\textit{effective fire temperature} of the pixel and to the
standard deviation $\sigma[T]$ as the associated
\textit{sub-pixel temperature spread}. The full pipeline runs at
flight cadence on the AVIRIS-3 on-board GPU.

\subsection{Identifying fire pixels for retrieval}
\label{qf}
\citet{dennison2009} found that empirical Hyperspectral Fire Detection
Index (HFDI) thresholds were capable of discriminating active fire from
the background across a wide range of fire temperatures, atmospheric water
vapor concentrations, and solar zenith angles. To identify candidate
fire pixels, we use HFDI, defined as:
\begin{equation}
\mathrm{HFDI}
=
\frac{L(2430\,\mathrm{nm}) - L(2061\,\mathrm{nm})}
     {L(2430\,\mathrm{nm}) + L(2061\,\mathrm{nm})},
\end{equation}
which captures the difference in spectral shape between reflected
solar radiance and fire-emitted radiance through the Planck function.
We apply a uniform threshold of $\mathrm{HFDI} \geq -0.1$ across all
scenes to select pixels for retrieval. In contrast,
\citet{matheson2012} selected the HFDI threshold dynamically per fire for images collected across a wider range of conditions (e.g., highly varying aircraft altitudes);
we illustrate how such selection could be adapted to this work in
\ref{figA1}. Many candidate pixels exhibit saturation in the SWIR
bands used for the HFDI calculation. These are retained for retrieval,
as saturation in these bands is itself a strong indicator of
high-temperature emission.

\subsection{Forward Model}
\label{fm}
\begin{figure}[H]
\centering
\includegraphics[width=\textwidth]{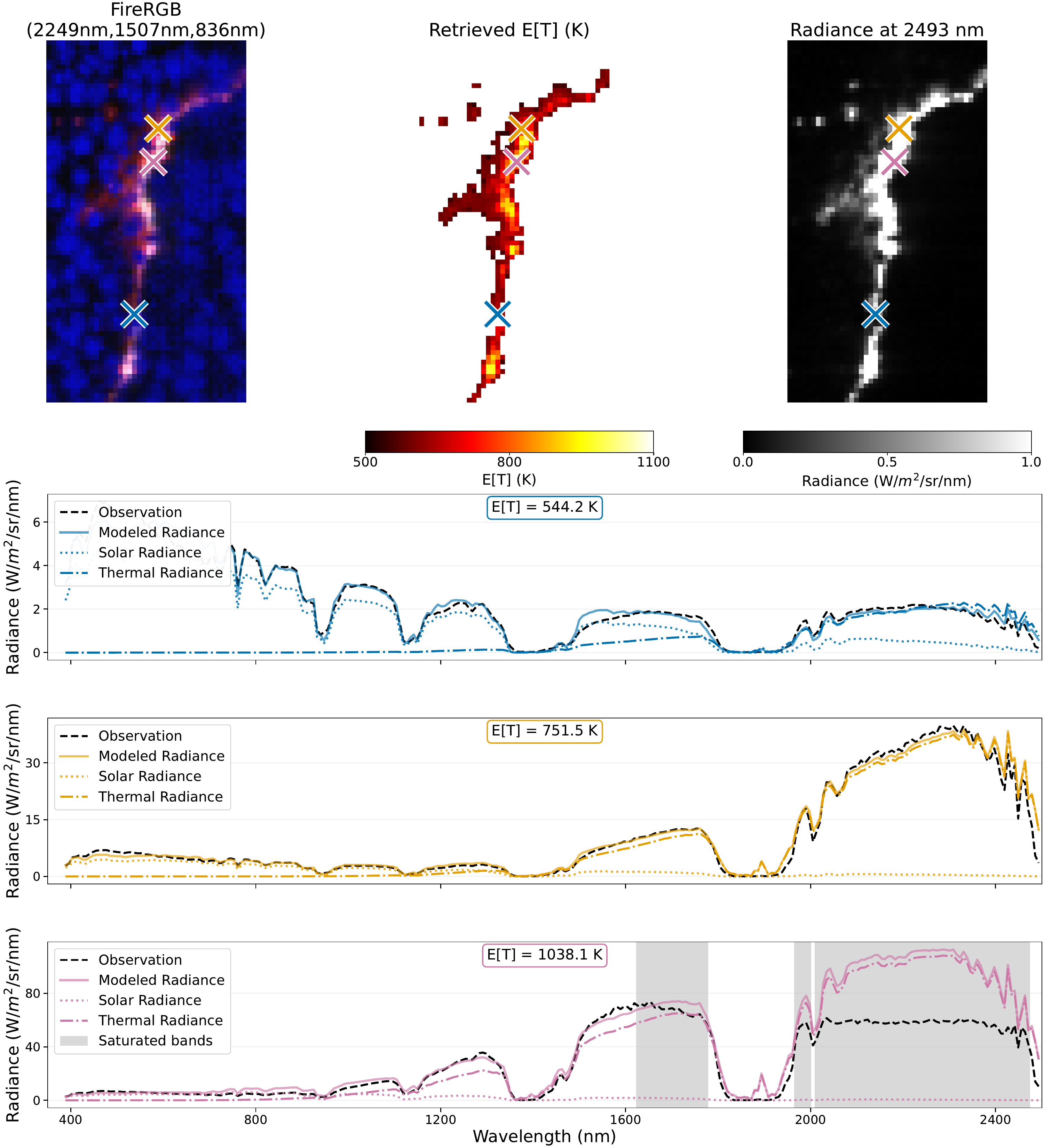}
\caption{Examples of observed and forward modeled spectrum from an AVIRIS-3 overflight of the Geneva Forest Fire (03/27/2025) for several retrieved effective fire temperatures $\mathbb{E}[T]$. "X" markers in the top row of images are centered on the pixels represented by the corresponding colored spectral plots below. Forward model radiance (solid color) is broken into the solar component (dotted) and thermal component (dash-dot). Saturated wavelengths are marked by the gray fill-between.}
\label{fig1}
\end{figure}

We model the observed radiance spectrum as the sum of a
solar-reflected term and a fire emitted term:
\begin{equation}
L_{\mathrm{model}}(\lambda)
=
L_{\mathrm{s}}(\lambda)
+
L_{\mathrm{t}}(\lambda)
\label{eq:fm}
\end{equation}
where $\lambda$ is wavelength, $L_{\mathrm{s}}$ is the solar
radiance, $L_{\mathrm{t}}$ is the emitted thermal radiance.
We use a 6S-based radiative transfer parameterization for the solar
term:
\begin{equation}
L_{\mathrm{s}}(\lambda)
=
\frac{E_{\odot}(\lambda)\cos\theta_s}{\pi}
\left[
\rho_{\mathrm{atm}}(\lambda)
+
\frac{
\tau_{\downarrow}(\lambda)\tau_{\uparrow}(\lambda)\rho(\lambda)
}{
1-S(\lambda)\rho(\lambda)
}
\right]
\label{eq:solar}
\end{equation}
where $E_{\odot}(\lambda)$ is solar irradiance, $\theta_s$ is solar
zenith angle, $\rho_{\mathrm{atm}}(\lambda)$ is atmospheric path
reflectance, $S(\lambda)$ is spherical sky albedo, and
$\tau_{\downarrow}(\lambda)$ and $\tau_{\uparrow}(\lambda)$ are the
downwelling and upwelling atmospheric transmittances, respectively.
All atmospheric terms are evaluated from a per-scene 6S
\citep{vermote1997} look-up table generated for a grid of column water-vapor values, following the look-up
table construction strategy as described in
\citet{thompson2018isofit} and \citet{brodrick2021srtmnet}.
The surface reflectance $\rho(\lambda)$ is represented as,
\begin{equation}
\rho(\lambda)
=
\sum_{k=1}^{5} x_{\mathrm{surf},k}\,\rho_k(\lambda),
\qquad
\sum_{k=1}^{5} x_{\mathrm{surf},k} = 1,
\qquad
x_{\mathrm{surf},k} \ge 0
\end{equation}
where $x_{\mathrm{surf},k}$ are the surface mixture fractions and
$\rho_k(\lambda)$ are a library of surface reflectance endmembers (Sec.~\ref{sec:lib}).
The emitted radiance term is constructed by integrating Planck radiance over
a temperature distribution:
\begin{equation}
L_{\mathrm{t}}(\lambda)
=
\alpha\,\tau_{\uparrow}(\lambda)
\int p(T)\,B(\lambda,T)\,dT
\label{eq:thermal}
\end{equation}
where $B(\lambda,T)$ is the blackbody emissive radiance at
temperature $T$ and $p(T)$ is the per-pixel temperature probability
density function. In practice the integral is evaluated as a discrete
sum over a temperature grid spanning $[350, 2200]\,\mathrm{K}$. A
retrieved scaling term $\alpha$ accounts for unknown gray-body
emissivity in $[0,1]$. We parameterize the sub-pixel temperature
distribution $p(T)$ as a moment-matched Gaussian obtained from a
two-component mixture. The hot (flaming) component is described by its
mean $\mu_{\mathrm{hot}}$ and standard deviation $\sigma_{\mathrm{hot}}$,
and the cooler (smoldering) component by its mean $\mu_{\mathrm{smol}}$
and standard deviation $\sigma_{\mathrm{smol}}$. Each component is
retrieved over its own SWIR inversion window,
$\Omega_{2}^{\mathrm{hot}}\approx[900,1800]\,\mathrm{nm}$ for the
flaming component and
$\Omega_{2}^{\mathrm{smol}}\approx[1500,2500]\,\mathrm{nm}$ for the
smoldering component, allowing the retrieval to adaptively select band
regions in the presence of saturation. Collapsing the mixture to a
single Gaussian by matching its first two moments and combining the
components with mixing weight $\eta\in[0,1]$, the moments of $p(T)$
(expected value and variance) are
\begin{equation}
\mathbb{E}[T]
=
\eta\,\mu_{\mathrm{hot}}
+(1-\eta)\,\mu_{\mathrm{smol}}
\end{equation}
\begin{equation}
\mathrm{Var}[T]
=
\eta\,\sigma_{\mathrm{hot}}^{2}
+(1-\eta)\,\sigma_{\mathrm{smol}}^{2}
+\eta(1-\eta)\,(\mu_{\mathrm{hot}}-\mu_{\mathrm{smol}})^{2}.
\end{equation}
We refer to the moment-matched mean $\mathbb{E}[T]$ throughout as the
\textit{effective fire temperature} of the pixel, with a standard
deviation $\sigma[T]=\sqrt{\mathrm{Var}[T]}$ representing the associated
sub-pixel temperature range. Fig.~\ref{fig1} illustrates the forward
model on three pixels drawn from the Geneva State Forest overflight of
March 27, 2025, selected to span the range of $\mathbb{E}[T]$
encountered in the campaign, showing the observed radiance alongside
the modeled radiance and its decomposition into solar and thermal
components.

\subsection{Pre-solved water vapor}

Column water vapor is solved prior to the temperature
retrieval using the depth of the 940 nm water absorption feature
relative to a continuum interpolated across the band shoulders. The
per-pixel water-vapor estimate is smoothed by replacing each pixel
with the median of its 12$\times$12 super-pixel neighborhood,
producing a spatially coherent atmospheric field (an example is
shown in the top right panel of Fig.~\ref{fig2}) that is then held
fixed during the temperature inversion. This decoupling is justified
by the spectral separation of the 940 nm water feature from the SWIR
windows that carry the thermal signal, and it converts what would
otherwise be a per-pixel atmospheric inversion into a small set of
quantized 6S evaluations cached for the scene.

\subsection{Spectral endmember library}
\label{sec:lib}

The surface reflectance basis $\{\rho_k(\lambda)\}$ is drawn from a
library of five fire-relevant endmembers (green vegetation,
non-photosynthetic vegetation, two soils, and a combined
char/ash class) sampled from the 2024 Lake Fire in Santa Barbara County, California \citep{ochoa_2025_17992843}. These spectra are resampled onto the AVIRIS-3 spectral grid using a
Gaussian spectral response function. The mixture is constrained to the unit simplex during optimization.
This library is not intended to be global or exhaustive, but to provide
enough plausible variation of background reflectance to facilitate reasonable
radiance fits while keeping the number of free parameters (in the fractions
of each endmember) small.

\subsection{Inversion}
\label{sec:inversion}

The forward model is inverted by minimizing a windowed least-squares
loss between modeled and observed radiance:
\begin{equation}
\mathcal{L}(\mathbf{x})
=
\sum_{\lambda \in \Omega}
m(\lambda)\,
\bigl[
L_{\mathrm{obs}}(\lambda) - L_{\mathrm{model}}(\lambda;\mathbf{x})
\bigr]^{2}
\label{eq:loss}
\end{equation}
where $\mathbf{x}$ is the per-pixel state vector,
$\Omega$ is a fitting window, and $m(\lambda)\in\{0,1\}$ is a
per-pixel band mask that excludes saturated bands (Sec.~\ref{methods}).
The optimization proceeds in two stages with different state
vectors and different fitting windows.

Before each optimization step, every continuous state variable
$x_{i}$ with physical bounds $[x_{i}^{\mathrm{lo}},
x_{i}^{\mathrm{hi}}]$ is mapped to a normalized coordinate
$z_{i}\in[-1,1]$ by the affine transform
\begin{equation}
z_{i}
=
\frac{2(x_{i} - x_{i}^{\mathrm{lo}})}{
x_{i}^{\mathrm{hi}} - x_{i}^{\mathrm{lo}}}
- 1,
\qquad
x_{i}
=
x_{i}^{\mathrm{lo}}
+ \frac{1}{2}(x_{i}^{\mathrm{hi}}-x_{i}^{\mathrm{lo}})(z_{i}+1).
\label{eq:affine}
\end{equation}
Optimization is performed in $z$-space, which equalizes the
effective Adam learning rate across variables with very different
physical scales (temperature in Kelvin, $\alpha$ unitless,
surface fractions on the unit simplex). The surface mixture
$\mathbf{x}_{\mathrm{surf}}=\{x_{\mathrm{surf},1},\ldots,
x_{\mathrm{surf},K}\}$ is additionally projected onto the unit
simplex $\{x_{\mathrm{surf},k}\ge 0,\
\sum_{k} x_{\mathrm{surf},k}=1\}$ after each step.

We use the Adam optimizer \citep{kingma2014} with running
estimates of the first and second moments of the gradient,
\begin{align}
\mathbf{m}_{t}
&=
\beta_{1}\,\mathbf{m}_{t-1}
+ (1-\beta_{1})\,\mathbf{g}_{t},
\\
\mathbf{v}_{t}
&=
\beta_{2}\,\mathbf{v}_{t-1}
+ (1-\beta_{2})\,\mathbf{g}_{t}^{2},
\\
\hat{\mathbf{m}}_{t}
&=
\mathbf{m}_{t}/(1-\beta_{1}^{t}),
\qquad
\hat{\mathbf{v}}_{t}
=
\mathbf{v}_{t}/(1-\beta_{2}^{t}),
\\
\mathbf{z}_{t}
&=
\mathbf{z}_{t-1}
- \mathrm{lr}\,
\frac{\hat{\mathbf{m}}_{t}}{\sqrt{\hat{\mathbf{v}}_{t}} + \epsilon},
\label{eq:adam}
\end{align}
where $\mathbf{g}_{t} =
\nabla_{\mathbf{z}}\mathcal{L}(\mathbf{x}(\mathbf{z}_{t-1}))$
is the loss gradient with respect to the normalized state, and
$\mathbf{m}_{t}$ and $\mathbf{v}_{t}$ are the running first and
second moments. We use the standard defaults
$\beta_{1}=0.9$, $\beta_{2}=0.999$, and $\epsilon=10^{-8}$, and clip
the gradient by the global norm to a maximum value of
$\|\mathbf{g}\|_{2}$ to prevent the saturated-band edge cases
from destabilizing the trajectory. After each update,
$\mathbf{z}_{t}$ is clipped back into $[-1,1]$ to enforce the
physical bounds.

The optimization is decomposed into two stages with different
state vectors and different fitting windows
$\Omega_{\mathrm{stage}}$, exploiting the spectral separation
between the solar and thermal contributions to the radiance.
\textit{Stage 1} fits the surface mixture coefficients
$\mathbf{x}_{\mathrm{surf}}$ together with a single scalar temperature
value $T_{1}$ and initial thermal scaling $\alpha_{1}$ using the
fitting window $\Omega_{1}=[350,2500]\,\mathrm{nm}$ where the solar
term carries the surface signal:
\begin{equation}
\mathbf{x}_{1}
=
\{x_{\mathrm{surf},1},\ldots,x_{\mathrm{surf},K},\ \alpha_{1},\ T_{1}\}.
\end{equation}
Stage 1 runs for 30 Adam steps at learning rate $\mathrm{lr}=0.03$.
\textit{Stage 2} fixes the surface mixture from Stage 1 and fits the
two-component Gaussian temperature mixture together with the
thermal scaling,
\begin{equation}
\mathbf{x}_{2}
=
\{\mu_{\mathrm{hot}},\sigma_{\mathrm{hot}},
\mu_{\mathrm{smol}},\sigma_{\mathrm{smol}},
\eta,\alpha\},
\end{equation}
over the SWIR fitting windows $\Omega_{2}^{\mathrm{hot}}$ and
$\Omega_{2}^{\mathrm{smol}}$ defined above, where the thermal term
dominates. Stage 2 runs for 100 Adam steps at $\mathrm{lr}=0.015$.
The smaller Stage 2 step size and longer horizon reflect the higher
conditioning of the thermal subproblem, where the Planck spectral
shape provides weaker per-step gradient signal than the surface
mixture does in Stage 1.

\begin{figure}[H]
\centering
\includegraphics[width=\textwidth]{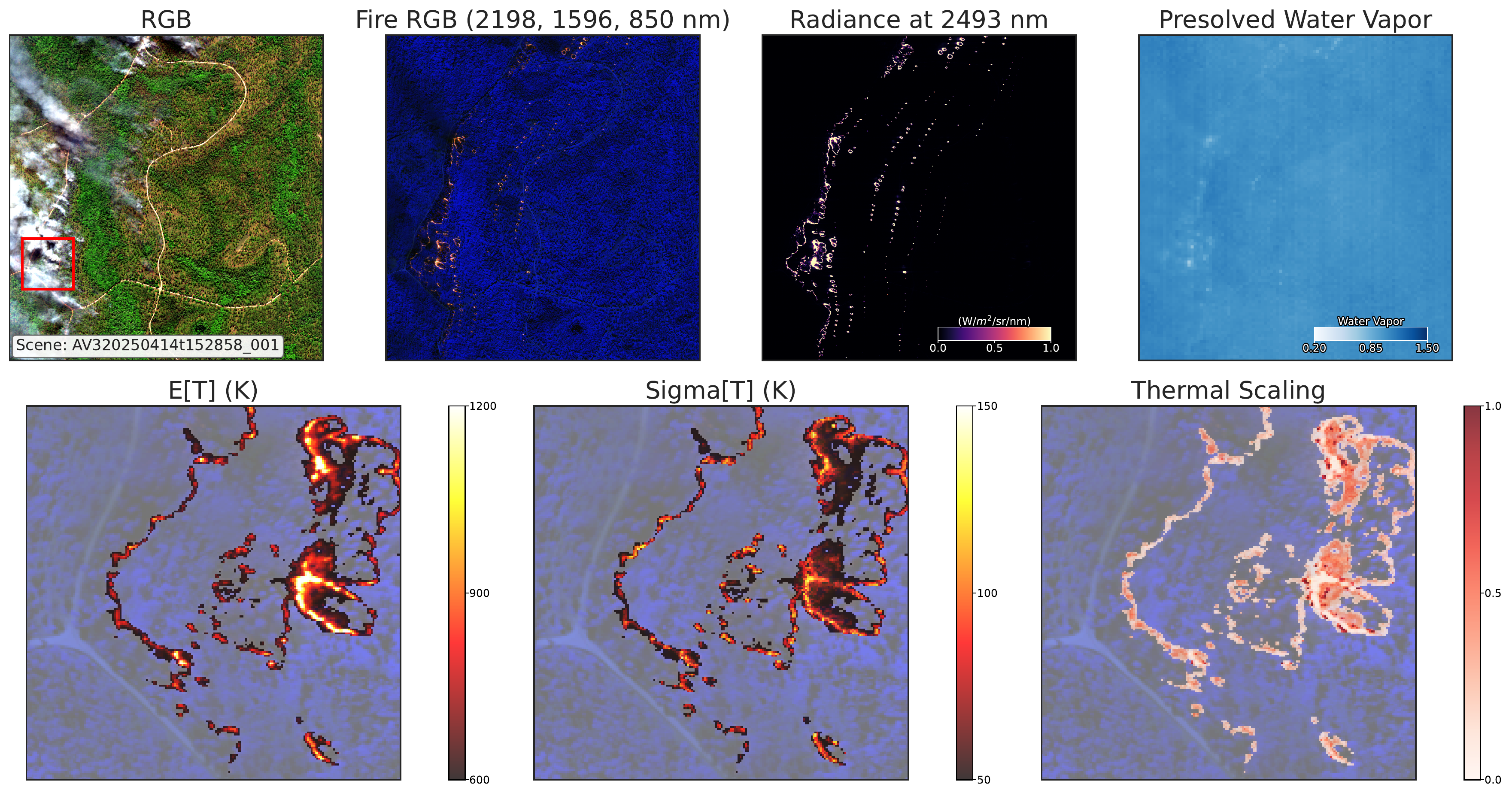}
\caption{Example of retrieved state variables for AVIRIS-3 scene AV320250111. Top row presolved quantities such as False Color RGB, the last band radiance used to identify active fire pixels, and the presolved superpixel water vapor. Bottom row: Shows the retrieved parameters from the moment matched Gaussian posterior and the retrieved thermal scaling, for a zoomed window from the top row (red box in the RGB).}
\label{fig2}
\end{figure}

The full inversion is implemented in JAX \citep{jax2018} and
batched over pixels, with per-scene 6S look-up tables built once
and evaluated for each pixel at its locally quantized water-vapor
value. Fig.~\ref{fig2} shows the full set of pre-solved and retrieved
state variables for one Fort Stewart scene
to illustrate the per-pixel outputs of
the pipeline. The pre-solved quantities (top row) are a true-color
RGB for context, a FireRGB composite (2198, 1596, 850 nm) that
highlights the active fire perimeter, the 2493 nm radiance which can indicate potential fire pixels, and the smoothed
super-pixel water vapor field. The retrieved quantities (bottom
row) are the effective fire temperature $\mathbb{E}[T]$, its
associated uncertainty $\sigma[T]$, and the thermal scaling
$\alpha$ from the moment-matched Gaussian posterior.

\subsection{Data}

We apply the retrieval to all AVIRIS-3 flight lines \citep{eckert2024aviris3_l1b} acquired during
the 2025 FireSense campaign over nine days (Fig.~\ref{fig4}, top),
spanning the Eaton Fire (Los Angeles, 11 January 2025), the Geneva
State Forest prescribed burn (Alabama, 27 March 2025), the Targets of
Collaboration wildfires and prescribed fires (Alabama, Florida, and Mississippi, over multiple days in March 2025), the
Crabapple Fire (Texas, 18 March 2025), and the Fort Stewart
prescribed burns (Georgia, April 2025). In total this dataset
comprises $\sim$168 scenes and $\sim$4 million plausible-fire pixels with an average spatial resolution of $\sim$5 m. Per-pixel
geometry, elevation, and sensor altitude are read from the per-scene
observation and location metadata, and the per-scene geometric lookup table is used to orthorectify the
retrieved temperature and uncertainty maps to a common geographic
grid. Further information on the FireSense campaign and its
associated data products is available at
\url{https://www.earthdata.nasa.gov/data/projects/firesense}.

\subsection{Emissive radiance injection into non-fire pixels}
To verify the inversion under controlled conditions we constructed
synthetic spectra by injecting a known single-temperature thermal
emission into measured AVIRIS-3 spectra drawn from non-fire pixels in
the same scenes. Rather than integrating over a temperature
distribution as in Eq.~\ref{eq:thermal}, each injection uses a single
temperature $T_{\mathrm{inj}}$, so the emitted term reduces to a
single Planck radiance:
\begin{equation}
L_{\mathrm{inj}}(\lambda)
=
L_{\mathrm{obs}}(\lambda)
+
\alpha_{\mathrm{inj}}\,\tau_{\uparrow}(\lambda)\,
B(\lambda, T_{\mathrm{inj}}),
\label{eq:inject}
\end{equation}
where $L_{\mathrm{obs}}(\lambda)$ is the observed non-fire radiance,
$T_{\mathrm{inj}}$ is the injected temperature drawn uniformly from
$[500, 1700]\,\mathrm{K}$, and $\alpha_{\mathrm{inj}}$ is a gray-body
emissivity scaling drawn uniformly at random. Each synthetic spectrum
was then passed through the full retrieval pipeline, and the retrieved
effective fire temperature $\mathbb{E}[T]$ was compared to the
injected temperature $T_{\mathrm{inj}}$.

\subsection{Spatial resolution rescaling}

To probe the behavior of the retrieval as the sensor footprint
coarsens, mimicking the situation a spaceborne instrument such as
EMIT would see relative to AVIRIS-3, we
aggregate fine $\sim$5 m AVIRIS-3 radiance to a coarse $\sim$60 m
grid (12$\times$12 block-mean in radiance space) and run the
retrieval independently at both resolutions, following the
fine-to-coarse aggregation approach of \citet{matheson2012}. The
fine-resolution retrievals within each coarse footprint are then
treated as an empirical per-pixel temperature distribution against
which the coarse-pixel moment-matched Gaussian posterior can be
evaluated. We summarize systematic departures between the two
across pixels with a Quantile Error Curve (QEC):
\begin{equation}
\mathrm{QEC}(q)
=
Q_{\mathrm{coarse}}(q) - Q_{\mathrm{fine}}(q),
\qquad
q\in\{0.05,0.10,0.25,0.50,0.75,0.90,0.95\}
\end{equation}
where $Q_{\mathrm{coarse}}$ is the quantile of the coarse Gaussian
posterior and $Q_{\mathrm{fine}}$ is the corresponding quantile of
either the empirical fine distribution or its Gaussian-mixture
reconstruction. Fifty scenes were sampled across the FireSense
sites for this analysis.

\section{Results}
\label{results}

The following results demonstrate the proposed retrieval in
multiple ways. First, we verified the forward model assumptions by
recovering a known single-temperature injection added to measured
background pixels with random thermal scaling. We then
characterized forward model residuals at the campaign scale across
the full FireSense 2025 dataset. Next, we evaluated the
consistency of the retrieval across spatial resolutions, the
adequacy of the Gaussian temperature-distribution assumption, and
the calibration of the retrieved uncertainty by aggregating fine
AVIRIS-3 pixels to coarser footprints representative of spaceborne
instruments and comparing the coarse posterior to the empirical
distribution of the underlying fine retrievals. Finally, we
summarize the population-scale distribution of retrieved effective
fire temperatures across the campaign.

\subsection{Forward model verification on injected background}
\label{sec:injection}

\begin{figure}[H]
\centering
\includegraphics[width=0.7\textwidth]{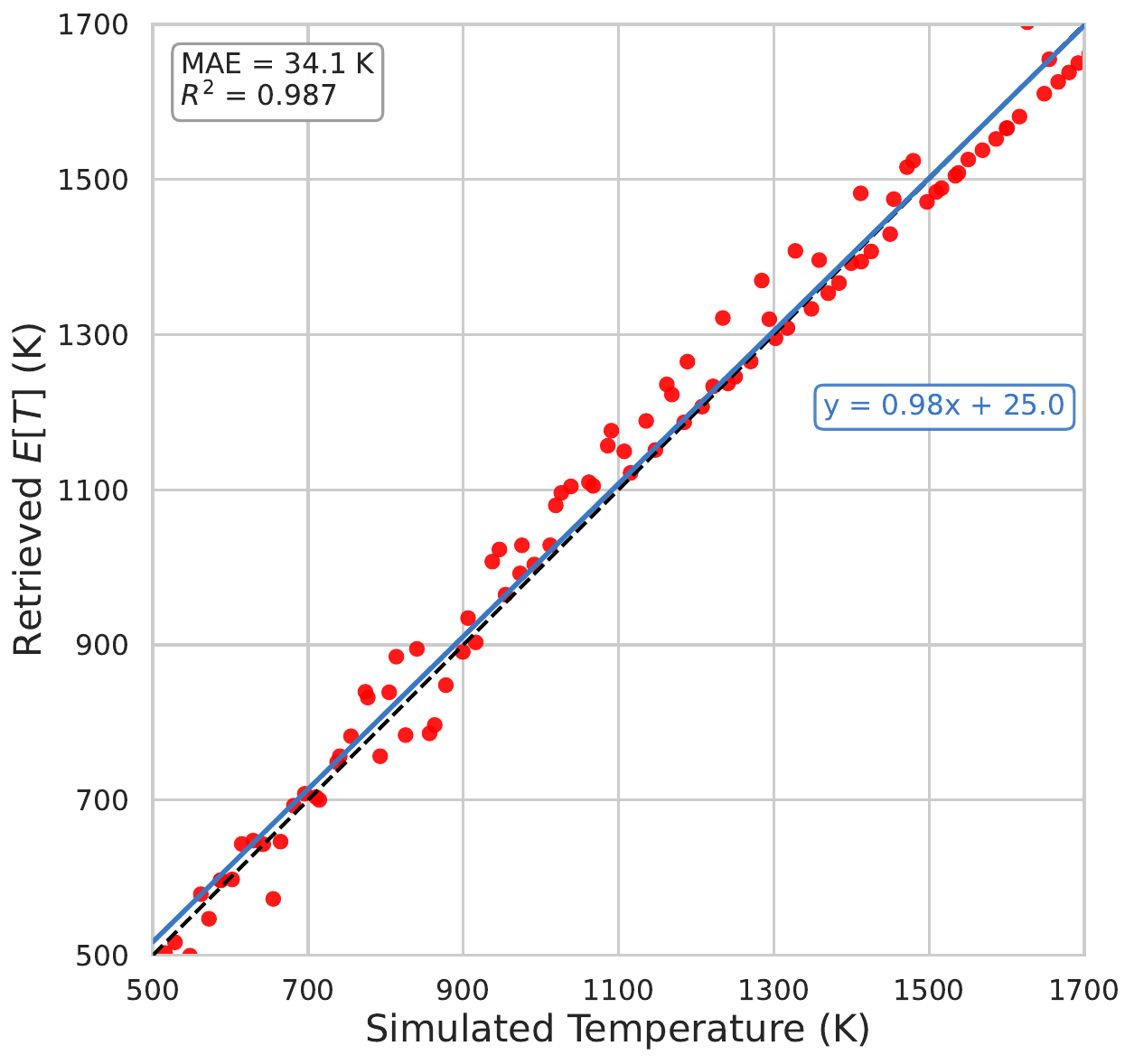}
\caption{Verifying retrieved temperature against simulation. We show the retrieved effective fire temperature $\mathbb{E}[T]$ versus the simulated temperature from the non-fire pixel emissivity injection.}
\label{fig3}
\end{figure}

We first verified that the inversion recovers a known thermal
signature when added to a measured background. Synthetic spectra
were constructed by injecting single-temperature Planck radiance
with a uniformly random thermal scaling into measured AVIRIS-3
background pixels (Sec.~\ref{methods}), spanning the 500-1700 K
range observed in the campaign. The retrieved effective fire
temperature $\mathbb{E}[T]$ from the full inversion is compared to
the injected temperature in Fig.~\ref{fig3}.

Over the 500-1700 K range the retrieval yields RMSE = 41.8 K, MAE = 34.1 K, and $R^{2}=0.987$. Scatter about the one-to-one
line is approximately symmetric across the range except for a tight clustering below the one-to-one line for the most extreme temperatures.

\subsection{Spectral residuals}
\label{sec:campaign}

\begin{figure}[H]
\centering
\includegraphics[width=\textwidth]{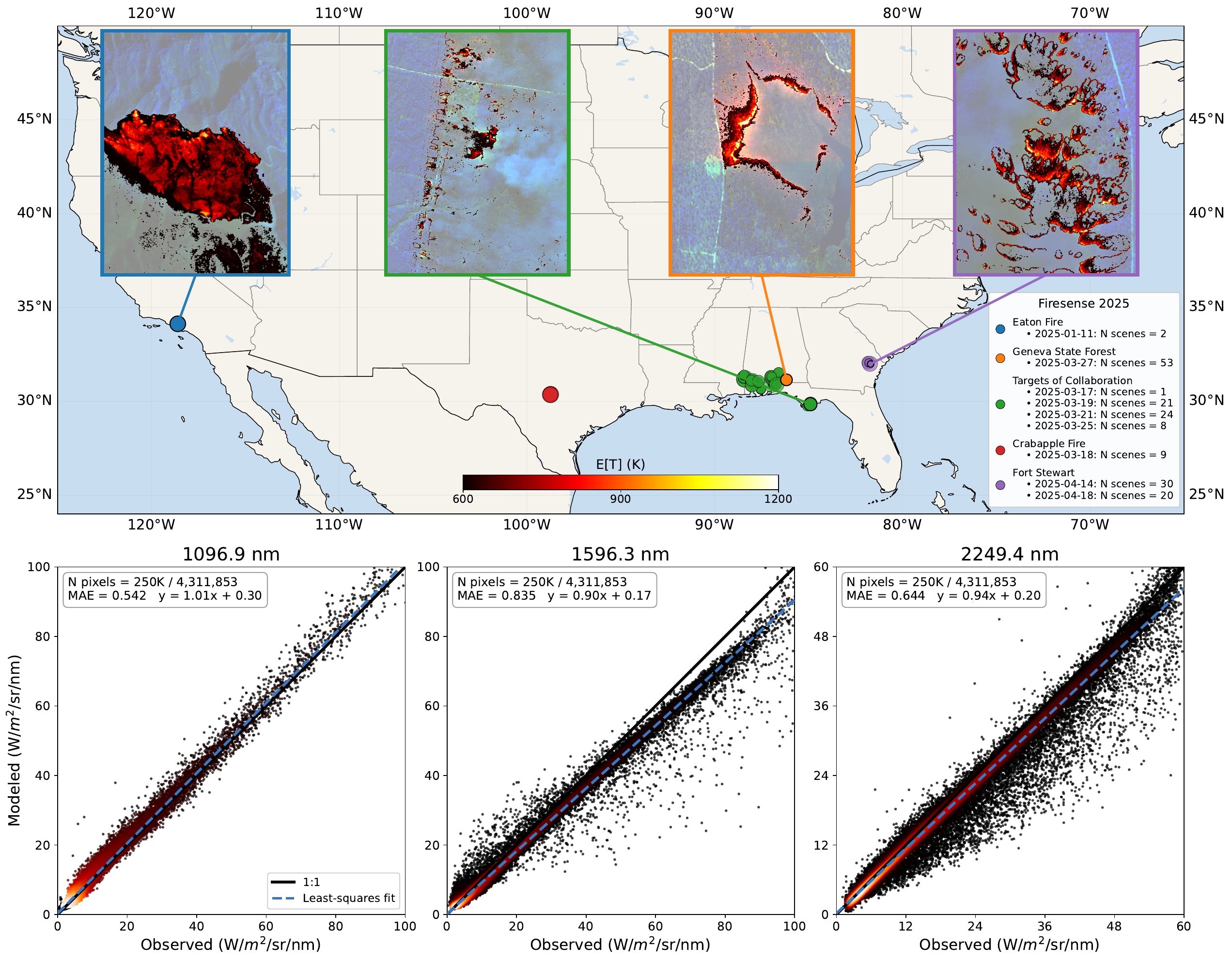}
\caption{Top: Firesense campaign locations used and examples of retrieved effective fire temperature $\mathbb{E}[T]$ (pixels not retrieved show FireRGB composite described in Section 2.4). Bottom: the residual between the forward modeled radiance and observation in the thermal window range at 1096.9 nm, 1596.3 nm and 2249.4 nm for pixels randomly sampled across FireSense 2025 sites - saturated points are excluded.}
\label{fig4}
\end{figure}

We next assess the forward model fit across the full FireSense 2025
dataset, comprising $\sim$168 scenes acquired over nine flight days
at six sites distributed from southern California to the Atlantic
coast (Fig.~\ref{fig4}, top). After applying the HFDI threshold,
$\sim$4 million plausible-fire pixels were retrieved campaign-wide.
To characterize the forward model fit on this dataset we randomly
sampled $\sim$250,000 pixels and compared the forward modeled
radiance to the observation at three diagnostic wavelengths chosen
to lie in spectral regions sensitive to thermal emission: 1097 nm, 1596 nm, and 2249 nm. Pixels with saturation in either band of the
pair were excluded from plotting.

At each of the three wavelengths the scatter clusters along the
one-to-one line with $R^{2}=0.98$ (Fig.~\ref{fig4}, bottom). Mean
residuals are $+0.35$ W\,m$^{-2}$\,sr$^{-1}$\,nm$^{-1}$ at 1097 nm,
$-0.67$ at 1596 nm, and $-0.55$ at 2249 nm, with standard
deviations of $0.87$, $1.52$, and $1.31$
W\,m$^{-2}$\,sr$^{-1}$\,nm$^{-1}$ respectively. The residual
spread is largest in the SWIR1 valley near 1596 nm for large radiance magnitudes.

\subsection{Robustness to spatial rescaling}
\label{sec:xres}

To probe retrieval behavior under the coarser footprints anticipated
for spaceborne instruments, and to
test both the adequacy of the Gaussian temperature-distribution
assumption and the calibration of the retrieved uncertainty, we
aggregated fine $\sim$5 m AVIRIS-3 radiance to a coarse $\sim$60 m
grid by 12$\times$12 block-mean averaging and ran the retrieval
independently at both resolutions for fifty scenes sampled across
the campaign. For each coarse pixel the fine-resolution retrievals
within its footprint give an empirical per-pixel distribution of
$\mathbb{E}[T]$ values, against which the coarse moment-matched
Gaussian posterior can be compared.

\begin{figure}[H]
\centering
\includegraphics[width=\textwidth]{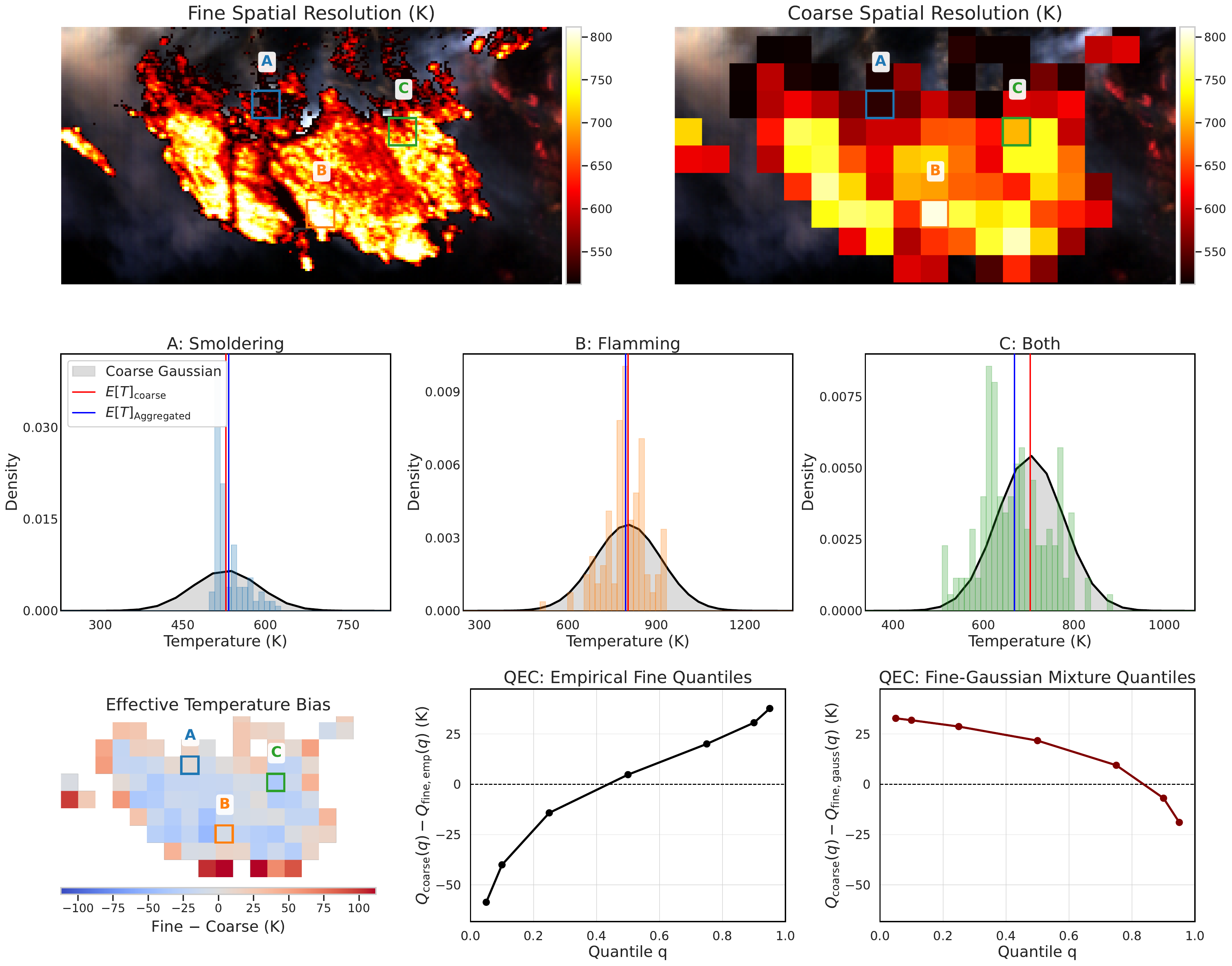}
\caption{Top row: retrieved effective fire temperature $\mathbb{E}[T]$ for active fire pixels at fine spatial resolution ($\sim$5~m$^2$) and at coarse spatial resolution ($\sim$60~m$^2$), from one of the 50 sampled scenes. Middle row: PDF comparison between the empirical aggregate distribution of fine-resolution $\mathbb{E}[T]$ within each coarse footprint (color histogram) and the coarse-pixel moment-matched Gaussian posterior (black curve) for the three annotated pixels. Bottom row: bottom left shows the per-pixel difference between the mean of the fine-resolution $\mathbb{E}[T]$ within a coarse footprint and the coarse-pixel $\mathbb{E}[T]$; bottom middle shows the average Quantile Error Curve (QEC) computed at the (5th, 10th, 25th, 50th, 75th, 90th, 95th) quantiles using the empirical fine-resolution $\mathbb{E}[T]$ distribution (black); bottom right shows the same QEC computed using the full Gaussian-mixture reconstruction of the fine-resolution posteriors (maroon).}
\label{fig5}
\end{figure}

The fine- and coarse-resolution $\mathbb{E}[T]$ maps for one of the
fifty scenes are shown in Fig.~\ref{fig5} (top row). Three coarse
pixels are annotated to span the regimes encountered: a smoldering
footprint (A, fine-pixel mean $\sim$530 K), a flaming footprint
(B, $\sim$800 K), and a mixed flaming and smoldering footprint (C,
$\sim$700 K). For each of these pixels the coarse moment-matched
Gaussian posterior is plotted over the histogram of the underlying
fine-resolution $\mathbb{E}[T]$ values (Fig.~\ref{fig5}, middle
row). 

To summarize the comparison across the full set of coarse pixels in
this scene, Fig.~\ref{fig5} (bottom left) shows the per-pixel
difference between $\mathbb{E}[T]_{\mathrm{coarse}}$ and the mean
of the fine-resolution $\mathbb{E}[T]$ values within the same
footprint. The difference is small and centered near zero across
most pixels, with positive and negative excursions at the fire-edge
pixels. The empirical Quantile Error Curve
(Fig.~\ref{fig5}, bottom middle) is monotonically increasing in
$q$ and crosses zero near $q=0.5$. At the 5th-10th quantiles the
coarse posterior is $\sim$50-75 K colder than the empirical fine
distribution, and at the 90th-95th quantiles it is $\sim$25-40 K
hotter. When the fine distribution is replaced by a Gaussian
mixture of its constituent posteriors (Fig.~\ref{fig5}, bottom
right), the Quantile Error Curve sign reverses and stays within
$\pm$30 K across all sampled quantiles.

\subsection{Campaign-scale temperature distribution}
\label{sec:population}

\begin{figure}[H]
\centering
\includegraphics[width=\textwidth]{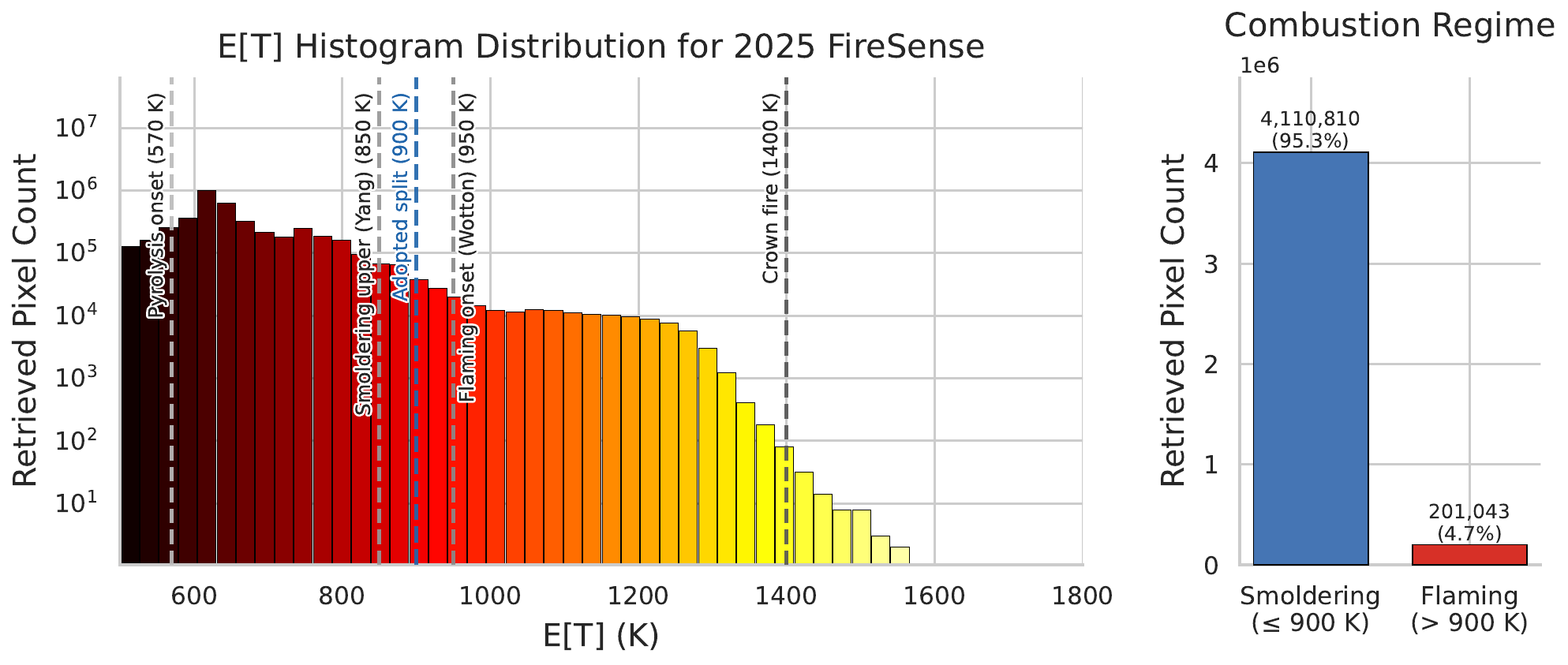}
\caption{Distribution of retrieved effective fire temperature
$\mathbb{E}[T]$ across the more than four million plausible-fire pixels
retrieved from the 2025 FireSense campaign, shown as binned pixel count
on a logarithmic axis (left). Gray dashed lines mark the pyrolysis and
volatile-release onset ($\sim$570\,K), the reported smoldering upper
bound ($\sim$850\,K; \citealp{yang2026forward}), the flaming onset
($\sim$950\,K; \citealp{wotton2012}), and crown fire ($\sim$1400\,K);
the blue dashed line marks the adopted smoldering/flaming split at
$900\,\mathrm{K}$. Right: retrieved pixel counts partitioned at
$900\,\mathrm{K}$ into smoldering ($\le900\,\mathrm{K}$;
$4{,}110{,}810$ pixels, $95.3\%$) and flaming ($>900\,\mathrm{K}$;
$201{,}043$ pixels, $4.7\%$) regimes.}
\label{fig6}
\end{figure}

We examine the distribution of retrieved effective fire temperatures
across the more than four million plausible-fire pixels from the 2025
FireSense campaign. Plotted on a log-count axis (Fig.~\ref{fig6}), the distribution appears to somewhat approximate a log-linear relation over much of its range, with a modest departure in the
 $1000$-$1400\,\mathrm{K}$ range. We note this structure
but do not model the falloff parametrically; instead we interpret the
distribution in terms of the two combustion regimes separated into a lower-temperature
smoldering regime and a higher-temperature flaming regime. Reported
smoldering effective temperatures extend up to roughly $850\,\mathrm{K}$
($\sim$577\,$^{\circ}$C) \citep{yang2026forward}, while flaming
combustion in forest fuels begins near $950\,\mathrm{K}$
($\sim$677\,$^{\circ}$C) \citep{wotton2012}. These bounds, also drawn in
Fig.~\ref{fig6}, leave a narrow transition band between them, and we
adopt $\mathbb{E}[T]=900\,\mathrm{K}$, at its midpoint, as the split
between smoldering-dominated and flaming-dominated pixels. For
reference, Fig.~\ref{fig6} marks reference temperatures for pyrolysis and volatile-release
onset ($\sim$570\,K; \citealp{white2001}) and crown fire
($\sim$1400\,K; \citealp{butler2004}).

Under this split, smoldering effective temperatures dominate the
campaign: $4{,}110{,}810$ pixels ($95.3\%$) fall in the smoldering
regime ($\mathbb{E}[T]\le900\,\mathrm{K}$) against $201{,}043$
($4.7\%$) in the flaming regime ($\mathbb{E}[T]>900\,\mathrm{K}$), so
pixels with a smoldering effective temperature are more than twenty
times as common as pixels with a flaming effective temperature
(Fig.~\ref{fig6}).

\begin{figure}[H]
\centering
\includegraphics[width=\textwidth]{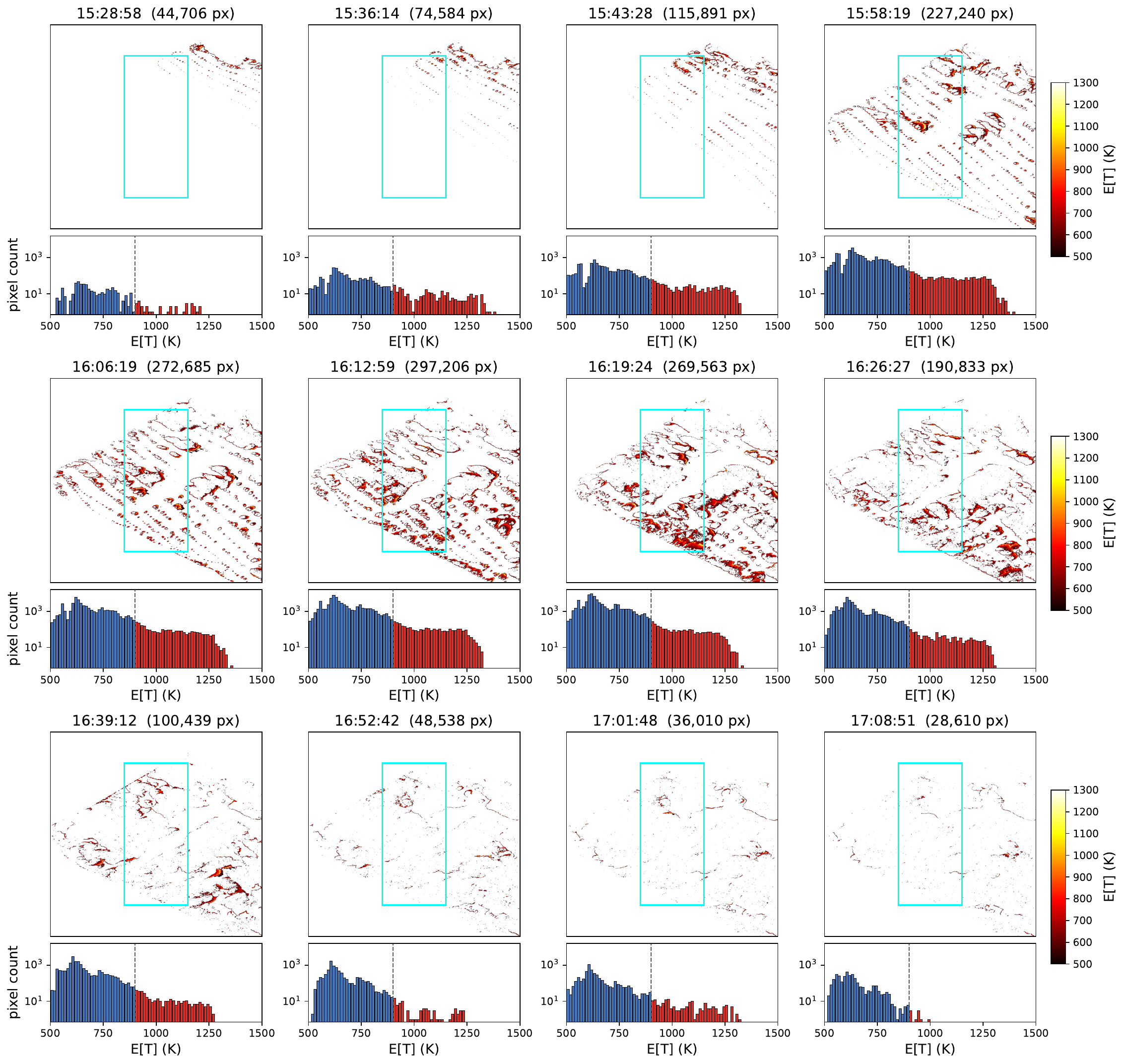}
\caption{Time series of retrieved effective fire temperature $\mathbb{E}[T]$
over the Fort Stewart overflight of 14 April 2025, with approximately
7 minutes between successive scenes. For each scene, the upper panel
shows the spatial map of $\mathbb{E}[T]$ (annotated with acquisition time
and in-region plausible-fire pixel count); the cyan box marks a fixed
collocated region common to all scenes, and the lower panel shows the
histogram of $\mathbb{E}[T]$ within that region, with the smoldering
($\le900\,\mathrm{K}$, blue) and flaming ($>900\,\mathrm{K}$, red)
regimes separated by the dashed line at $900\,\mathrm{K}$. Flaming temperatures become less prevalent toward the end of the time series.}
\label{fig7}
\end{figure}

We also leverage the repeat VSWIR imaging spectroscopy that the 2025
FireSense campaign provides through multiple AVIRIS-3 overflights of
of prescribed fires, so the retrieved effective temperature field can be followed
as a time series rather than a single acquisition. The Fort Stewart
overflight of a prescribed fire ignited by delayed aerial ignition devices on April 14, 2025 illustrates this capability: it was
reacquired at an approximately 7-minute cadence, and Fig.~\ref{fig7}
shows the retrieved $\mathbb{E}[T]$ field for each of the twelve scenes.
To make the scenes directly comparable, the pixel counts and histograms
are computed over a fixed collocated region common to all scenes (cyan
box) rather than the full flight-line. Within this region, the
number of plausible-fire pixels rises from roughly $45{,}000$ in the
first scene to a peak near $297{,}000$ at 16:12:59 before declining to
roughly $29{,}000$ by 17:08:51. Every scene remains dominated by the
smoldering regime, with the flaming tail most pronounced in the
mid-sequence scenes of highest pixel count. The spatial maps show the
active-fire area expanding to fill the collocated region through the
middle of the sequence and contracting toward the end.

\section{Discussions and Conclusions}
\label{discussion}

We have demonstrated a
real-time capable, physics-based inversion for per-pixel wildfire
temperature retrieval from airborne imaging spectroscopy, applied
at campaign scale across $\sim$168 AVIRIS-3 flight lines and
$\sim$4 million plausible-fire pixels spanning the 2025
FireSense campaign. Because the retrieval inverts an explicit
radiative-transfer forward model rather than fitting a contextual
or band-ratio summary statistic, it absorbs effects that
point-estimate methods must pre-compute, assume away, or fold into
empirical correction factors. Specifically, it directly facilitates pixel-specific atmospheric and geometric radiative transfer  \citep{brodrick_isofit}, enables complex surface mixtures with endmember mixing, and includes an optimizer-resolved coupling of gray-body emissivity through the retrieved scaling term $\alpha$.
On simulated spectra with known injected thermal emissions, the
retrieval recovers the effective fire temperature $\mathbb{E}[T]$
over the 500-1700 K simulated range with a mean absolute error of
$34.1$ K (Fig.~\ref{fig3}), and the forward model
achieves close residual fit at three diagnostic SWIR wavelengths across
the $\sim$4 million plausible-fire pixels of the FireSense 2025
campaign (Fig.~\ref{fig4}). The forward model produces good
residual fits particularly in the wavelengths most sensitive to
thermal emission, and the two-stage Adam optimizer runs within the
flight time of a scene on the AVIRIS-3 on-board GPU, making the
retrieval suitable for operational deployment. Additionally, adaptively selecting which bands influence the fit through mixture density gives the retrieval robustness in highly saturated regions for very hot pixels. 

Previous spectral-mixture retrievals from imaging spectrometer data
represent per-pixel temperature heterogeneity through a small set
of discrete blackbody emitted radiance endmembers \citet{dennison2006,dennison2011comparison,matheson2012,waigl2019,amici2022}. These
retrievals return the component temperature(s) and their mixing
fractions as point estimates without propagated per-pixel
uncertainty. Our forward model replaces this discrete
representation with a continuous Gaussian parameterization of the
per-pixel temperature population and reports it as a moment-matched
single Gaussian per pixel, exposing both the effective fire
temperature $\mathbb{E}[T]$ and the associated uncertainty
$\sigma[T]$ in a single retrieval. Across the coarse-resolution
validation set (Fig.~\ref{fig5}), the moment-matched Gaussian
posterior at coarse resolution agrees with the empirical
distribution of underlying fine-spatial resolution $\mathbb{E}[T]$
values to within a calibration spread of approximately 95 K
between the 5th and 95th quantiles, with the largest departures in
the lower tail at smoldering pixels near the minimum detectable
thermal signal of the AVIRIS-3 instrument. We interpret the
lower-tail departure not as a bias of the retrieval but as the
posterior correctly absorbing the uncertainty about unobservable
per-pixel temperatures into a wider variance. The mean difference
between the average of fine-resolution $\mathbb{E}[T]$ values
within a footprint and the coarse-resolution $\mathbb{E}[T]$
retrieved for that same footprint exhibits an RMSE of $27.16$ K,
indicating that the retrieval is robust to changes in instrument
footprint, a property that will matter when porting the algorithm
from AVIRIS-3 ($\sim$5 m) to EMIT \citep{green2020emit}
($\sim$60 m) and EAGLE-VSWIR ($\sim$30 m).

Aggregating $\mathbb{E}[T]$ across the more than four million
plausible-fire pixels in the FireSense 2025 campaign
(Fig.~\ref{fig6}), smoldering effective temperatures outnumber flaming
ones by more than an order of magnitude. This regime balance is not
static: because combustion evolves from flaming toward smoldering as
active fronts pass and fuels enter residual burning, and because
smoldering can persist far longer than flaming, the relative balance of
the two regimes over a fire is inherently a temporal quantity. A single
acquisition captures only one moment of that evolution. The repeat
AVIRIS-3 overflights in the campaign provide the temporal component
directly: the Fort Stewart time series (Fig.~\ref{fig7}) follows the
retrieved $\mathbb{E}[T]$ field of a single fire at an approximately
7-minute cadence, resolving how the smoldering and flaming populations
shift as the fire develops and decays. The relative abundance of smoldering vs. flaming combustion from the Fort Stewart time series should not be assumed to be representative for other fires, since this time series captures a prescribed fire with over a thousand ignition points. Extending this analysis across
the multiple fires imaged in the campaign would allow the
flaming-to-smoldering transition and the persistence of smoldering to be
characterized at population scale rather than inferred from a single
snapshot.

Several limitations apply to the present approach. First, the
retrieval assumes a Gaussian temperature distribution at each
pixel. This is best understood as a plausible representation of the potential fire temperature distributions. The SWIR
radiance signature constrains the mean and width of the underlying
temperature population, but it does not in general carry enough
shape information to distinguish a true Gaussian from a multimodal
mixture or from a uniform ramp between two endpoints (determined by experimentation of the forward model, results not shown). Reporting
the moment-matched single Gaussian is therefore not a claim about
the true per-pixel temperature distribution, but a calibrated
summary of what the spectrum constrains. Second, gray-body
emissivity is absorbed into the thermal scaling $\alpha$ rather
than retrieved independently; depending on the imaging conditions, the optimizer may leverage
$\alpha$ to absorb the fractional emitting area of the
pixel (the quantity the classical retrievals try to
isolate) and adjacency effects from neighboring bright thermal
pixels \citep{dennison2006}. A partial $\alpha$-temperature degeneracy is also quite possibly
present. The spectral amplitude of the thermal contribution can
in principle be reproduced either by a larger $\alpha$ at a lower
mean temperature or by a smaller $\alpha$ at a higher mean
temperature, with the Planck spectral shape providing the only
leverage to separate the two. This trade-off has been shown to be
ill-posed for SWIR-only retrievals at low thermal contrast
\citep{dennison2006,giglio2008}, and although we have not
quantified its effect in the present retrieval, it may contribute
to the larger absolute departures observed at the lower end of
the injection-recovery range (Sec.~\ref{sec:injection}). These
degeneracies will remain until an independent emissivity or area
constraint is brought in. Third, the forward model assumes the surface
reflectance can be expressed as a sparse mixture of fire-relevant
endmembers; pixels containing unusual surface materials (e.g.,
burned structures in wildland-urban interface fires such as the
Eaton Fire) may exhibit elevated forward model residual. Additionally, contemporaneous sensor artifacts, such as nonlinearity, may continue to pose challenges, as seen in the increase in residual error in the SWIR1 shoulder
near 1600 nm. This is a limitation shared by all SWIR-based fire retrievals
\citep{dennison2006,giglio2008} and one motivation for a richer
posterior representation rather than a point estimate at the
saturated end.

Given the potential portability of the retrieval to other
imaging spectrometers, future work could readily expand the application
of this algorithm to orbital imaging spectrometers like EMIT. Collocation with
VIIRS active-fire detections \citep{schroeder2014,giglio2016}
could then provide a long temporal baseline and an independent
point-of-comparison from the operational thermal-infrared product,
enabling assessment of detection consistency across the EMIT
record. The co-located ECOSTRESS land surface temperature record
\citep{fisher2020ecostress}, acquired from the same ISS platform,
additionally offers thermal-infrared coverage of the regime below
the $\sim$500 K floor at which the SWIR signal in EMIT becomes
detectable, providing complementary temperatures in the post-combustion
range that the present approach cannot retrieve directly. Also worth
investigating are forward model discrepancies, possibly addressed
through a stricter surface-reflectance prior, and the use of repeat
overflights to follow the combustion-regime balance over time, linking
the retrieved-temperature time series quantitatively to pre-fire fuel
maps \citep{andrews2014} and weather conditions. Operationally, the
on-board $\mathbb{E}[T]$ and $\sigma[T]$ products are well suited to
direct ingestion into fire-spread models \citep{mandel2014} and
emissions estimates \citep{ichoku2014}, where the calibrated uncertainty
enables formal data assimilation rather than the deterministic insertion
currently used for thermal-infrared products.

\section{Declaration of Competing Interest}
\label{competinginterest}
The authors declare no competing interest.

\section{Data and Code Availability Statement}
\label{dataandcode}
AVIRIS-3 data for this work are publicly available and free to download via the AVIRIS-3 L1B facility instrument collection \citep{eckert2024aviris3_l1b}. Spectra collected from the 2024 Lake Fire are available for download and use a native resolution \citep{ochoa_2025_17992843}. Code will be made publicly available at \url{https://github.com/isofit/isofire} upon publication. 

\section{Acknowledgments}
\label{acknowledgments}
We thank the members of the AVIRIS-3 flight team and FireSense collaborators who participated in data acquisition and analysis. AVIRIS-3 is sponsored by the National Aeronautics and Space Administration (NASA) Earth Science Division. This research was carried out at the Jet Propulsion Laboratory, California Institute of Technology, under a contract with the National Aeronautics and Space Administration. A portion of this work was funded by NASA ROSES 80NM0018F0590 and by Realtime Active Fire Data from NASA's AVIRIS-3 Sensor for Fire Operations 80NSSC25K0225. Copyright 2021 California Institute of Technology. All rights reserved. US Government Support Acknowledged.

\appendix
\section{Interaction between HFDI threshold selection and solar angle}
\label{app1}
\begin{figure}[H]
\centering
\includegraphics[width=\textwidth]{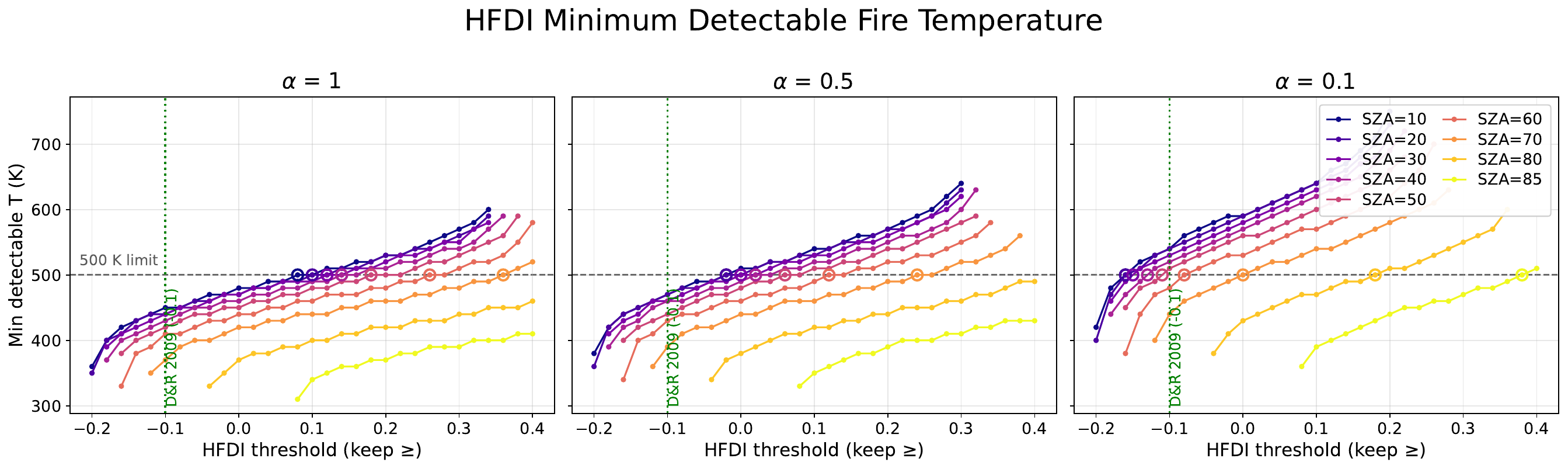}
\caption{Minimum detectable fire temperature as a function of the HFDI
detection threshold, for gray-body emissivity scalings $\alpha=1$, $0.5$,
and $0.1$ (panels, left to right) and solar zenith angles from $10$ to
$85^{\circ}$ (colored curves). Each curve gives the lowest forward-modeled
fire temperature whose $\mathrm{HFDI}=(L_{2430}-L_{2061})/(L_{2430}+L_{2061})$
meets or exceeds the threshold on the horizontal axis, at fixed surface
reflectance ($0.5$) and column water vapor ($1.8\,\mathrm{cm}$). The green
dotted line marks the literature threshold $\mathrm{HFDI}=-0.1$
\citep{dennison2009}; the dashed horizontal line marks a $500\,\mathrm{K}$
reference, and open circles mark where each curve crosses it. At fixed
threshold, higher SZA and larger $\alpha$ lower the minimum detectable
temperature.}
\label{figA1}
\end{figure}

The uniform $\mathrm{HFDI}\ge-0.1$ threshold applied in Sec.~\ref{qf}
follows the value reported by \citet{dennison2009}, but the temperature
at which a sub-pixel fire becomes detectable by this test depends on both
solar illumination and the strength of the emitted thermal signal. To
characterize this dependence we forward-modeled the at-sensor radiance
with the same 6S-based model used in the retrieval (Sec.~\ref{fm}), with a column
water vapor at $1.8\,\mathrm{cm}$ and adding a single-temperature Planck
term scaled by the upwelling transmittance and a gray-body emissivity
scaling $\alpha$; for each solar zenith angle (SZA) and $\alpha$ we swept
the temperature from $300$ to $2000\,\mathrm{K}$, computed
$\mathrm{HFDI}=(L_{2430}-L_{2061})/(L_{2430}+L_{2061})$ at each step, and
took the minimum detectable temperature at a given threshold to be the
lowest fire temperature whose HFDI meets or exceeds it (Fig.~\ref{figA1}).
Increasing SZA lowers the minimum detectable temperature at a fixed
threshold because a longer solar path reduces the reflected-solar
background against which the thermal signal competes, while decreasing gray-body 
($\alpha$) weakens the emitted thermal contribution and requires a more
permissive threshold to reach the same minimum detectable temperature. A
fixed threshold therefore does not correspond to a fixed detection
temperature across a scene: the same $\mathrm{HFDI}\ge-0.1$ cut admits
different coldest-detectable fires depending on illumination and
sub-pixel thermal strength, which motivates the per-scene threshold
adaptation discussed in Sec.~\ref{qf}.

\bibliographystyle{elsarticle-harv}
\bibliography{references}

\end{document}